\documentclass[10pt,twocolumn,letterpaper]{article}
\usepackage{cvpr}
\usepackage{times}
\usepackage{epsfig}
\usepackage{graphicx}
\usepackage{amsmath}
\usepackage{amssymb}
\usepackage{subcaption}
\usepackage{algorithmicx,algorithm}
\usepackage[noend]{algpseudocode}
\newcommand\sbullet[1][.5]{\mathbin{\vcenter{\hbox{\scalebox{#1}{$\bullet$}}}}}

\usepackage[breaklinks=true,bookmarks=false]{hyperref}

\cvprfinalcopy 

\ifcvprfinal\fi
\begin{document}
\title{Incremental Learning In Online Scenario}
\author{Jiangpeng He\\
{\tt\small he416@purdue.edu}
\and
Runyu Mao\\
{\tt\small mao111@purdue.edu}
\and
Zeman Shao\\
{\tt\small shao112@purdue.edu}
\and
Fengqing Zhu\\
{\tt\small zhu0@purdue.edu}
\and
{School of Electrical and Computer Engineering, Purdue University, West Lafayette, Indiana USA}
}
\maketitle
\thispagestyle{empty}
\begin{abstract}
Modern deep learning approaches have achieved great success in many vision applications by training a model using all available task-specific data. However, there are two major obstacles making it challenging to implement for real life applications: (1) Learning new classes makes the trained model quickly forget old classes knowledge, which is referred to as catastrophic forgetting. (2) As new observations of old classes come sequentially over time, the distribution may change in unforeseen way, making the performance degrade dramatically on future data, which is referred to as concept drift. Current state-of-the-art incremental learning methods require a long time to train the model whenever new classes are added and none of them takes into consideration the new observations of old classes. In this paper, we propose an \textbf{incremental learning framework} that can work in the challenging online learning scenario and handle both new classes data and new observations of old classes. We address problem (1) in online mode by introducing a modified cross-distillation loss together with a two-step learning technique. Our method outperforms the results obtained from current state-of-the-art offline incremental learning methods on the CIFAR-100 and ImageNet-1000 (ILSVRC 2012) datasets under the same experiment protocol but in online scenario. We also provide a simple yet effective method to mitigate problem (2) by updating exemplar set using the feature of each new observation of old classes and demonstrate a real life application of online food image classification based on our complete framework using the Food-101 dataset. 
\end{abstract}

\section{Introduction}
\label{introduction}
One of the major challenges of current deep learning based methods when applied to real life applications is learning new classes incrementally, where new classes are continuously added overtime. Furthermore, in most real life scenarios, new data comes in sequentially, which may contain both the data from new classes or new observations of old classes. Therefore, a practical vision system is expected to handle the data streams containing both new and old classes, and to process data sequentially in an online learning mode~\cite{ONLINELEARNING}, which has similar constrains as in real life applications. For example, a food image recognition system designed to automate dietary assessment should be able to update using each new food image continually without forgetting the food categories already learned.

Most deep learning approaches trained on static datasets suffer from the following issues. First is catastrophic forgetting~\cite{CF}, a phenomenon where the performance on the old classes degrades dramatically as new classes are added due to the unavailability of the complete previous data.
This problem become more severe in online scenario due to limited run-time and data allowed to update the model. The second issue arises in real life application where the data distribution of already learned classes may change in unforeseen ways~\cite{data_distribution_in_real_life}, which is related to concept drift~\cite{CD}. 
In this work, we aim to develop an incremental learning framework that can be deployed in a variety of image classification problems and work in the challenging online learning scenario.

A practical deep learning method for classification is characterized by (1) its ability to be trained using data streams including both new classes data and new observations of old classes, (2) good performance for both new and old classes on future data streams, (3) short run-time to update with constrained resources, and (4) capable of lifelong learning to handle multiple classes in an incremental fashion. Although progress has been made towards reaching these goals~\cite{LWF,ICARL,EEIL,BiC}, none of the existing approaches for incremental learning satisfy all the above conditions.
They assume the distribution of old classes data remain unchanged overtime and consider only new classes data for incoming data streams. As we mentioned earlier, data distribution are likely to change in real life\cite{data_distribution_in_real_life}. When concept drift happens, regardless the effort put into retaining the old classes knowledge, degradation in performance is inevitable. 
In addition, although these existing methods have achieved state-of-the-art results, none of them work in the challenging online scenario. They require offline training using all available new data for many epochs, making it impractical for real life applications.

The main contributions of this paper is summarized as follows.
\begin{itemize}
  \item We introduce a modified cross-distillation loss together with a two-step learning technique to make incremental learning feasible in online scenario. We show comparable results to the current state-of-the-art~\cite{ICARL,EEIL,BiC} on CIFAR-100~\cite{CIFAR} and ImageNet-1000 (ILVSC2012)~\cite{IMAGENET1000}. We follow the same experiment benchmark protocol~\cite{ICARL} where all new data belong to new class, but in the challenging online learning scenario where the condition is more constrained for both run-time and number of data allowed to update the model.
  \item We propose an incremental learning framework that is capable of lifelong learning and can be applied to a variety of real life online image classification problems. In this case, we consider new data belong to both new class and existing class. We provide a simple yet effective method to mitigate concept drift by updating the exemplar set using the feature of each new observation of old classes. Finally, we demonstrate how our complete framework can be implemented for food image classification using the Food-101~\cite{Food-101} dataset.
\end{itemize}

\section{Related Work}
\label{relatedwork}
In this section, we review methods that are closely related to our work. Incremental learning remains one of the long-standing challenges for machine learning, yet it is very important to brain-like intelligence capable of continuously learning and knowledge accumulation through its lifetime.

\textbf{Traditional methods.} Prior to deep learning, SVM classifier~\cite{svm} is commonly used. One representative work is~\cite{svm1}, which learns the new decision boundary by using support vectors that are learned from old data together with new data. An alternative method is proposed in~\cite{svm2} by retaining the Karush-Kuhn-Tucker conditions instead of support vectors on old data and then update the solution using new data. Other techniques~\cite{polikar2001learn++,NCM2,kuzborskij2013n} use ensemble of weak classifiers and nearest neighbor classifier.

\textbf{Deep learning based methods.} These methods provide a joint learning of task-specific features and classifiers. Approaches such as~\cite{jung2016less,kirkpatrick2017overcoming} are based on constraining or freezing the weights in order to retain the old tasks performance. In~\cite{jung2016less}, the last fully connected layer is freezed which discourages change of shared parameters in the feature extraction layers. Inn \cite{kirkpatrick2017overcoming} old tasks knowledge is retained by constraining the weights that are related to these tasks. However, constraining or freezing parameters also limits its adaptability to learn from new data. A combination of knowledge distillation loss~\cite{kd} with standard cross-entropy loss is proposed to retain the old classes knowledge in~\cite{LWF}, where old and new classes are separated in multi-class learning and distillation is used to retain old classes performance. However,  performance is far from satisfactory when new classes are continuously added, particularly in the case when the new and old classes are closely related. Based on~\cite{LWF}, auto encoder is used to retain the knowledge for old classes instead of using distillation loss in \cite{rannen2017encoder}. For all these methods, only new data is considered.

In~\cite{shin2017continual} and \cite{venkatesan2017strategy}, synthetic data is used to retain the knowledge for old classes by applying a deep generative model~\cite{GAN}. However, the performance of these methods are highly dependent on the reliability of the generative model, which struggles in more complex scenarios. 

Rebuffi et al proposed iCaRL\cite{ICARL}, an approach using a small number of exemplars from each old class to retain knowledge. An end-to-end incremental learning framework is proposed in \cite{EEIL} using exemplar set as well, along with data augmentation and balanced fine-tuning to alleviate the imbalance between the old and new classes. Incremental learning for large datasets was proposed in \cite{BiC} in which a linear model is used to correct bias towards new classes in the fully connected layer. However, it is difficult to apply these methods to real life applications since they all require a long offline training time with many epochs at each incremental step to achieve a good performance. In addition, they assume the distribution of old classes remain unchanged and only update the classifiers using new classes data.
All in all, a modified cross-distillation loss along with a two-step learning technique is introduced to make incremental learning feasible in the challenging online learning scenario. Furthermore, our complete framework is capable of lifelong learning from scratch in online mode, which is illustrated in Section~\ref{Our Complete Method}.


\section{Online Incremental Learning}
\label{onlinelearning}
\begin{figure}[t]
\begin{center}
  \includegraphics[width=1.\linewidth]{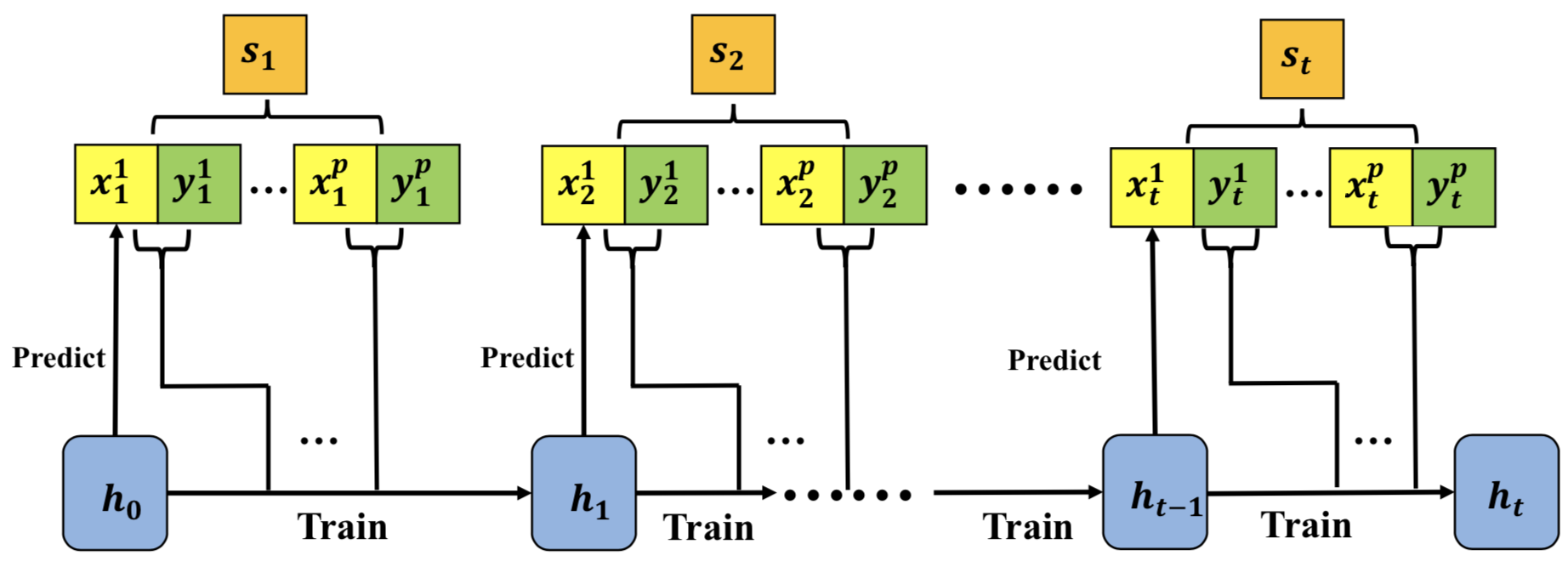}
  \caption{\textbf{Online Scenario.} A sequence of model $h_1, h_2,...,h_{t}$ is generated using each block of new data with block size $p$, where $(\textbf{x}^i_t,y^i_t)$ indicate the i-th new data for the t-th block.}
  \label{fig:1}
  \vspace{-.7cm}
\end{center}
\end{figure}

Online incremental learning~\cite{ONLINELEARNING} is a subarea of incremental learning that are additionally bounded by run-time and capability of lifelong learning with limited data compared to offline learning. However, these constraints are very much related to real life applications where new data comes in sequentially and is in conflict with the traditional assumption that complete data is available. A sequence of model $h_1, h_2,...,h_t$ is generated on the given stream of data blocks $s_1, s_2, ...,s_t$ as shown in Figure~\ref{fig:1}. In this case, $s_i$ is a block of new data with block size $p$, defined as the number of data used to update the model, which is similar to batch size as in offline learning mode. However, each new data is used only once to update the model instead of training the model using the new data with multiple epochs as in offline mode. $s_t = \{ (\textbf{x}_t^{(1)},y_t^{(1)}),...,(\textbf{x}_t^{(p)},y_t^{(p)})\}  \in R^n\times \{1,...,M\}$ where n is the data dimension and $M$ is the total number of classes. The model $h_t : R^n \rightarrow \{1,...,M\}$ depends solely on the model $h_{t-1}$ and the most recent block of new data $s_t$ consisting of $p$ examples with $p$ being strictly limited, e.g. if we set $p=16$ then we will predict for each new data and use a block of 16 new data to update the model. 

Catastrophic forgetting is the main challenge faced by all incremental learning algorithms. Suppose a model $h_{base}$ is initially trained on $n$ classes and we update it with $m$ new added classes to form the model $h_{new}$. Ideally, we hope $h_{new}$ can predict all $n+m$ classes well, but in practice the performance on the $n$ old classes drop dramatically due to the lack of old classes data when training the new classes. In this work, we propose a modified cross-distillation loss and a two-step learning technique to address this problem in online scenario.

Concept drift is another problem that happens in most real life applications. Concept~\cite{CONCEPT} in classification problems is defined as the joint distribution $P(X,Y)$ where $X$ is the input data and $Y$ represents target variable. Suppose a model is trained on data streams by time $t$ with joint distribution $P(X_t,Y_t)$, and let $P(X_n,Y_n)$ represent the joint distribution of old classes in future data streams. Concept drift happens when $P(X_t,Y_t) \neq P(X_n,Y_n)$. In this work, we do not measure concept drift quantitatively, but we provide a simple yet effective method to mitigate the problem by updating the exemplar set using the features of each new data in old classes, which is illustrated in Section~\ref{method_part3}


\section{Incremental Learning Framework}
\label{Our Complete Method}
\label{Method}
\begin{figure}[t]
\begin{center}
  \includegraphics[width=0.9\linewidth]{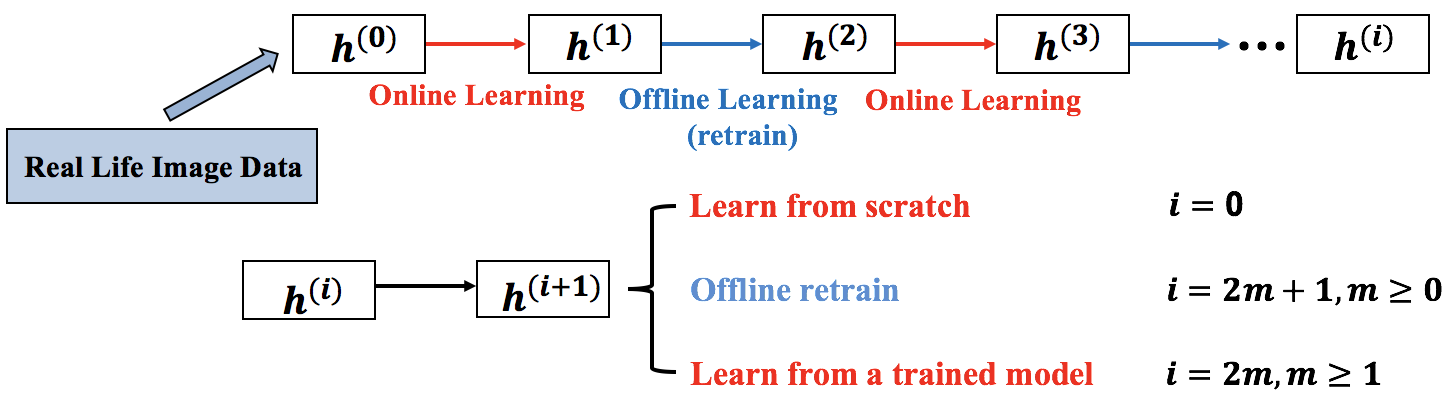}
  \caption{\textbf{Proposed incremental learning framework. } $h^{(i)}$ indicates the evolving model at i-th step.}
  \vspace{-.9cm}
  \label{fig:0}
\end{center}
\end{figure}

In this work, we propose an incremental learning framework as shown in Figure~\ref{fig:0} that can be applied to any online scenario where data is available sequentially and the network is capable of lifelong learning. There are three parts in our framework: \textit{learn from scratch}, \textit{offline retraining} and \textit{learn from a trained model}.  Incremental learning in online scenario is implemented in~\ref{method_part3} and lifelong learning can be achieved by alternating the last two parts after initial learning.
\subsection{Learn from Scratch}
\label{method_part1}
This part serves as the starting point to learn new classes. In this case, we assume the network does not have any previous knowledge of incoming classes, which means there is no previous knowledge to be retained. Our goal is to build a model that can adapt to new classes fast with limited data, e.g. block size of 8 or 16.

\textbf{Baseline.} 
Suppose we have data streams with block size $p$ belong to $M$ classes: $\{s_1,...,s_t\} \in R^n\times \{1,...,M\}$. The baseline for the model to learn from sequential data can be thought as generating a sequence of model $\{h_1,...,h_t\}$ using standard cross-entropy where $h_t$ is updated from $h_{t-1}$ by using block of new data $s_t$. Thus $h_t$ is evolving from $h_0$ for a total of $t$ updates by using the given data streams. Compared to traditional offline learning, the complete data is not available and we need to update the model for each block of new data to make it dynamically fit to the data distribution used so far. So in the beginning, the performance on incoming data is poor due to data scarcity. 

\textbf{Online representation learning.} 
A practical solution is to utilize representation learning when data is scarce at the beginning of the learning process. Nearest class Mean (NCM) classifier~\cite{NCM_IN, ICARL} is a good choice where the test image is classified as the class with the closest class data mean. We use a pre-trained deep network to extract features by adding a representation layer before the last fully connected layer for each input data $\textbf{x}_i$ denoted as $\phi(\textbf{x}_i)$. Thus the classifier can be expressed as
\begin{equation} 
\label{eq:1}
y^{\ast} = \mathop{\arg\min}_{y \in \{1,...,M\}} \ \  d (\phi(\textbf{x}),\mu_y^{\phi}).
\end{equation}
The class mean $\mu_y^{\phi} = \frac{1}{N_y}\sum_{i:y_i=i}\phi(\textbf{x}_i)$ and $N_y$ denote the number of data in classes $y$. We assume that the highly non-linear nature of deep representations eliminates the need of a linear metric and allows to use Euclidean distance here
\begin{equation} 
\label{eq:2}
d^{\phi}_{xy} = (\phi(\textbf{x}) - \mu_y^{\phi})^T(\phi(\textbf{x}) - \mu_y^{\phi})
\end{equation} 

\textbf{Our method: combining baseline with NCM classifier.} 
NCM classifier behaves well when number of available data is limited since the class representation is based solely on the mean representation of the images belonging to that class. We apply NCM in the beginning and update using an online estimate of the class mean~\cite{DEEPNCM} for each new observation.
\begin{equation} 
\label{eq:1}
\mu_y^{\phi} \leftarrow \frac{n_{yi}}{n_{yi}+1}\mu_y^{\phi} + \frac{1}{n_{yi}+1}\phi(\textbf{x}_i)
\end{equation}
We use a simple strategy to switch from NCM to baseline classifier when accuracy for baseline surpass representation learning for $s$ consecutive blocks of new data. Based on our empirical results, we set $s=5$ in this work.


\subsection{Offline Retraining}
\label{offline retraining}
In order to achieve lifelong learning, we include an offline retraining part after each online incremental learning phase. By adding new classes or new data of existing class, both catastrophic forgetting and concept drift~\cite{CD} become more severe. The simplest solution is to include a periodic offline retraining by using all available data up to this time instance. 

\textbf{Construct exemplar set.} We use herding selection~\cite{HERDING} to generate a sorted list of samples of one class based on the distance to the mean of that class. We then construct the exemplar set by using the first $q$ samples in each class $\{E_1^{(y)},...E_q^{(y)}\}, y \in [1,...,n]$ where $q$ is manually specified. The exemplar set is commonly used to help retain the old classes' knowledge in incremental learning methods.

\begin{figure}[H]
\begin{center}
  \includegraphics[width=0.9\linewidth]{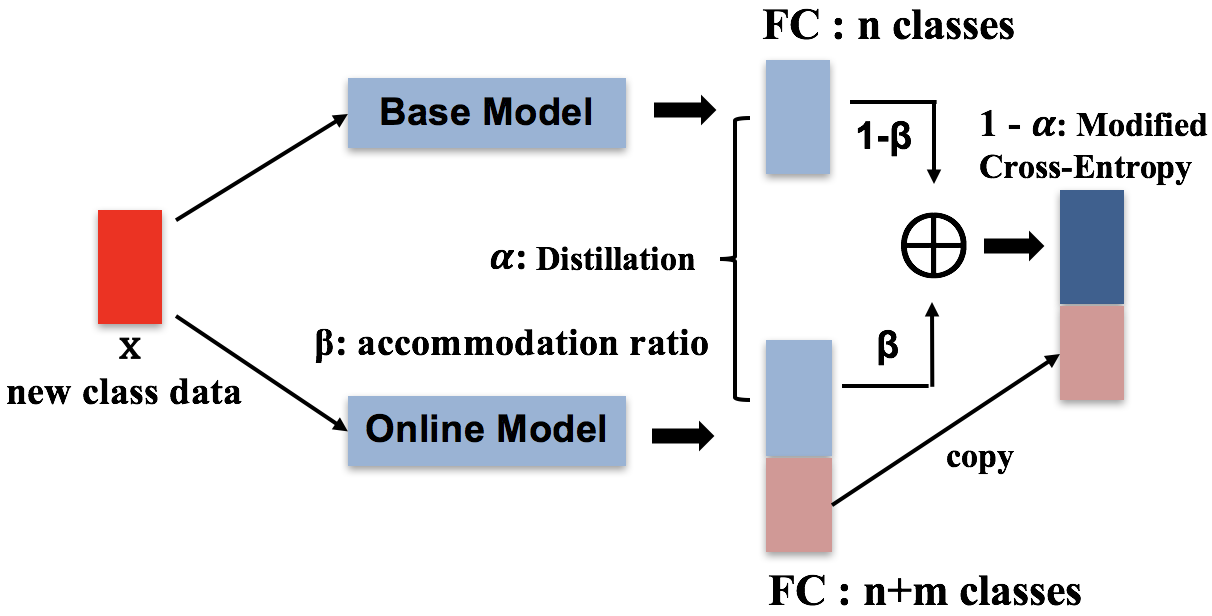}
  \vspace{-0.2cm}
  \caption{\textbf{Modified Cross-Distillation Loss.} It contains two losses: the distilling loss on old classes and the modified cross-entropy loss on all old and new classes.}
  \label{fig:3}
\end{center}
\end{figure}
\vspace{-.8cm}

\subsection{Learn from a Trained Model}
\label{method_part3}
This is the last component of our proposed incremental learning framework.  The goal here is to continue to learn from new data streams starting from a trained model. Different from existing incremental learning, we define new data containing both new classes data and new observations of old classes and we use each new data only once for training in online scenario. In additional to addressing the catastrophic forgetting problem, we also need to consider concept drift for already learned classes due to the fact that data distribution in real life application may change over time in unforeseen ways~\cite{data_distribution_in_real_life}.

\textbf{Baseline: original cross-distillation loss.} Cross-distillation loss function is commonly used in state-of-the-art incremental learning methods to retain the previous learned knowledge. In this case, we consider only new classes data for incoming data streams. Suppose the model is already trained on $n$ classes, and there are $m$ new classes added. Let $\{(\textbf{x}_i,y_i), y_i \in [n+1,...n+m]\}$ denote new classes data. The output logits of the new classifier is denoted as $p^{(n+m)}(x) = (o^{(1)},...,o^{(n)},o^{(n+1)}, ...o^{(n+m)})$, the recorded old classes classifier output logits is $\hat{p}^{(n)}(x) = (\hat{o}^{(1)},...,\hat{o}^{(n)})$. The knowledge distillation loss~\cite{kd} can be formulated as in Equation~\ref{eq:7}, where $\hat{p}_T^{(i)}$ and $p_T^{(i)}$ are the i-th distilled output logit as defined in Equation~\ref{eq:8}
\begin{equation} \label{eq:7}
\begin{aligned}
L_{D}(x) = \sum_{i=1}^n-\hat{p}_T^{(i)}(x)log[p_T^{(i)}(x)]
\end{aligned}
\end{equation}
\begin{equation} \label{eq:8}
\begin{aligned}
\hat{p}_T^{(i)} =
\frac{\exp{(\hat{o}^{(i)}}/T)}{\sum_{j=1}^n\exp{(\hat{o}^{(j)}}/T)}\ , \ 
p_T^{(i)} = \frac{\exp{(o^{(i)}}/T)}{\sum_{j=1}^n\exp{(o^{(j)}}/T)}
\end{aligned}
\end{equation}
$T$ is the temperature scalar. When $T= 1$, the class with the highest score has the most influence. When $T >1$, the remaining classes have a stronger influence, which forces the network to learn more fine grained knowledge from them. The cross entropy loss to learn new classes can be expressed as $L_{C}(x) = \sum_{i=1}^{n+m}-\hat{y}^{(i)}log[p^{(i)}(x)]$
where $\hat{y}$ is the one-hot label for input data $x$. 
The overall cross-distillation loss function is formed as in Equation~\ref{eq:10} by using a hyper-parameter $\alpha$ to tune the influence between two components. 
\begin{equation} \label{eq:10}
\begin{aligned}
L_{CD}(x) = \alpha L_D(x) + (1-\alpha) L_C(x)
\end{aligned}
\end{equation} 

\textbf{Modified cross-distillation with accommodation ratio.} Although cross-distillation loss forces the network to learn latent information from the distilled output logits, its ability to retain previous knowledge still remains limited. An intuitive way to make the network retain previous knowledge is to keep the output from the old classes' classifier as a part of the final classifier. Let output logits of the new classifier be denoted as $p^{(n+m)}(x) = (o^{(1)},...,o^{(n)},o^{(n+1)}, ...o^{(n+m)})$, the recorded old classes' classifier output logits is $\hat{p}^{(n)}(x) = (\hat{o}^{(1)},...,\hat{o}^{(n)})$. We use an accommodation ratio $0 \leq \beta \leq 1$ to combine the two classifier output as 
\begin{equation} \label{eq:11}
\Tilde{p}^{(i)}=\left\{
\begin{array}{rcl}
&\beta p^{(i)} + (1-\beta)\hat{p}^{(i)} &  {0<i\leq n}\\&p^{(i)} & {n < i\leq n+m}
\end{array} \right.
\end{equation}
When $\beta = 1$, the final output is the same as the new classifier and when $\beta = 0$, we replace the first $n$ output units with the old classes classifier output. This can be thought as using the accommodation ratio $\beta$ to tune the output units for old classes. As shown in Figure~\ref{fig:3}, the modified cross-distillation loss can be expressed by replacing the original cross-entropy loss part $L_C(x)$ with the new modified cross-entropy loss $\Tilde{L}_{C}(x) = \sum_{i=1}^{n+m}-\hat{y}^{(i)}log[\Tilde{p}^{(i)}(x)]$ after applying the accommodation ratio as in Equation~\ref{eq:modified cd loss}
\begin{equation} \label{eq:modified cd loss}
\begin{aligned}
\Tilde{L}_{CD}(x) = \alpha L_D(x) + (1-\alpha) \Tilde{L}_{C}(x)
\end{aligned}
\end{equation} 
We empirically set $\beta = 0.5$, $T=2$ and $\alpha = \frac{n}{n+m}$ in this work where $n$ and $m$ are the number of old and new classes.
The modified cross-distillation loss push the network to learn more from old classes' output units since we add it directly as part of the final output.
\begin{algorithm}[t]
\caption{Update Exemplar Set}
\hspace*{0.02in} {\bf Input:} 
New observation for old classes $(\textbf{x}_i, y_i)$\\
\hspace*{0.02in} {\bf Require:} 
Old classes feature extractor \textbf{$\Theta$}\\
\hspace*{0.02in} {\bf Require:} 
Current exemplar set $\{E_1^{(y_i)},...E_q^{(y_i)}\}$
\begin{algorithmic}[1]
\State$M^{(y_i)} \leftarrow \frac{n_{y_i}}{n_{y_i}+1}M^{(y_i)} + \frac{1}{n_{y_i}+1}\textbf{$\Theta$}(\textbf{x}_i)$
\For{m = 1,...,q}
\State $d^{(m)} = (\Theta(E_m^{(y_i)}) - M^{(y_i)})^T(\Theta(E_m^{(y_i)}) - M^{(y_i)})$
\EndFor\\
$d_{min} \leftarrow \min\{d^{(1)},...,d^{(m)}\}$\\
$I_{min} \leftarrow \textrm{Index}\{d_{min}\}$\\
$d^{(q+1)} = (\Theta(\textbf{x}_i) - M^{(y_i)})^T(\Theta(\textbf{x}_i) - M^{(y_i)})$
\If{$d^{(q+1)} \leq d_{min}$}
\State Remove $E_{I_{min}}^{(y_i)}$ from $\{E_1^{(y_i)},...E_q^{(y_i)}\}$
\State Add $x_i$ to $\{E_1^{(y_i)},...E_{q-1}^{(y_i)}\}$
\Else
\State No need to update current exemplars
\EndIf
\State \Return $\{E_1^{(y_i)},...E_q^{(y_i)}\}$
\end{algorithmic}
\label{alg:update exemplar set}
\end{algorithm} 

\textbf{Update exemplar set.} 
As described in Section~\ref{introduction}, we consider the new data streams containing both new classes data and new observations of old classes with unknown distribution. In this case, retaining previous knowledge is not sufficient since concept drift can happen to old classes and the model will still undergo performance degradation. One solution is to keep updating the network using the exemplars for old classes. The class mean of each old class $\{M^{(1)},...,M^{(n)}, \ M^{(i)} \in R^n \}$ is calculated and recorded as described in Section~\ref{offline retraining} by constructing the exemplar set $\{(E_1^{(y)},...E_q^{(y)}), y \in [1,...,n]\}$ using previous data streams. Let $\{(\textbf{x}_i, y_i), y_i \in [1,...,n]\}$ denote the new observation of old classes. We follow the same online class mean update as described in Equation~\ref{eq:1} to keep updating the class mean with all data seen so far. So when concept drift happens, e.g., the class mean changes toward the new data, we update the exemplar set to make new data more likely to be selected to update the model during two-step learning as described in next part. The complete process of updating exemplar set is shown in Algorithm~\ref{alg:update exemplar set}. 

\begin{figure}[H]
\begin{center}
  \includegraphics[width=1.\linewidth]{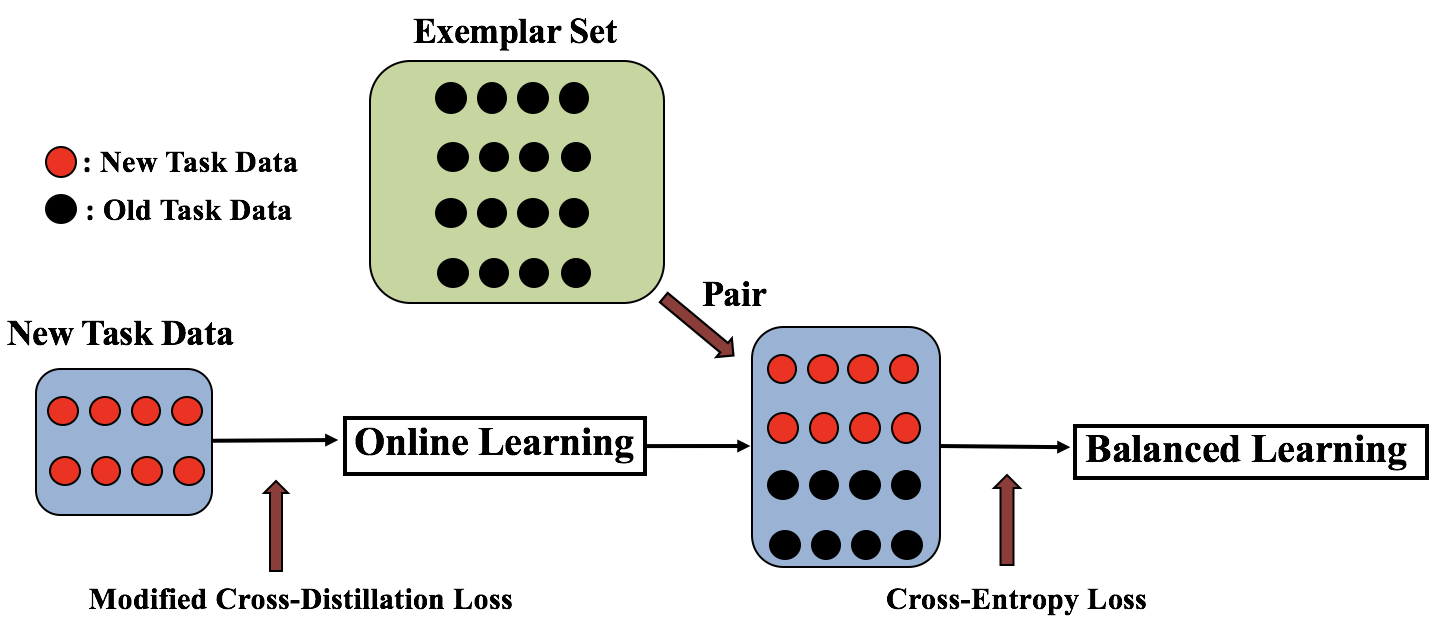}
  \vspace{-0.6cm}
  \caption{\textbf{Two-Step Learning.} Black dots correspond to old classes samples stored in exemplar set. Red dots correspond to samples from new classes.}
  \vspace{-0.6cm}
  \label{fig:2step}
\end{center}
\end{figure}

\textbf{Two-step learning.} Unlike other incremental learning algorithms that mix new classes data with old classes exemplars, we first let the model learn from a block of new classes data and then a balanced learning step is followed. This two-step learning technique is deigned for online learning scenarios, where both update time and number of available data are limited. As shown in Figure~\ref{fig:2step}, we use the modified cross-distillation loss in the first step to overcome catastrophic forgetting since all data in this block belongs to new classes. In the second step, we pair same number of old classes exemplars from the exemplar set with the new classes data. As we have balanced new and old classes, cross entropy loss is used to achieve balanced learning.

\begin{figure*}[htbp]
\centering
\begin{minipage}[t]{0.22\linewidth}
    \centering
    \includegraphics[width=4.cm,height=3.5cm]{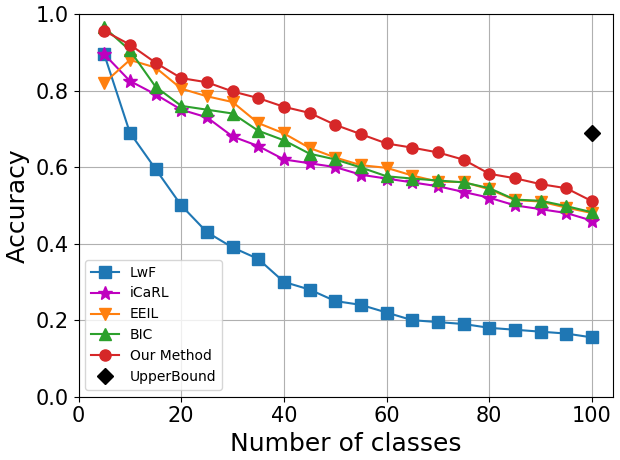}
    \parbox{15.5cm}{\small \hspace{2.cm}(a)}
\end{minipage}
\hspace{2ex}
\begin{minipage}[t]{0.22\linewidth}
    \centering
    \includegraphics[width=4.cm,height=3.5cm]{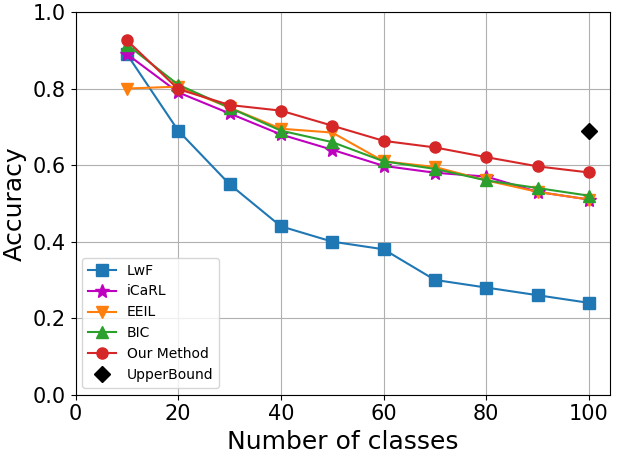}
    \parbox{15.5cm}{\small \hspace{2.cm}(b)}
\end{minipage}
\hspace{2ex}
\begin{minipage}[t]{0.22\linewidth}
    \centering
    \includegraphics[width=4.cm,height=3.5cm]{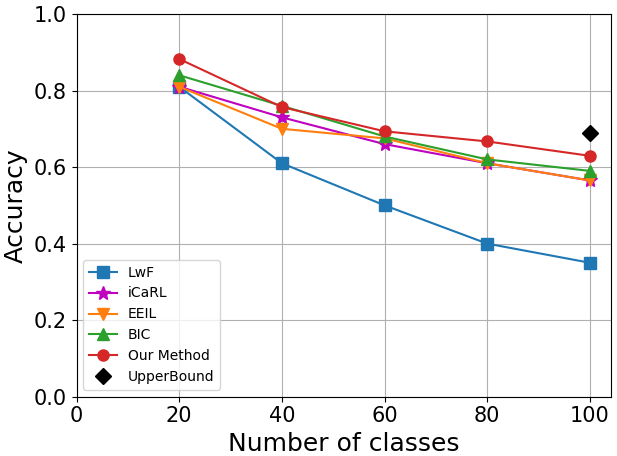}
    \parbox{15.5cm}{\small \hspace{2.cm}(c)}
\end{minipage}
\hspace{2ex}
\begin{minipage}[t]{0.22\linewidth}
    \centering
    \includegraphics[width=4.cm,height=3.5cm]{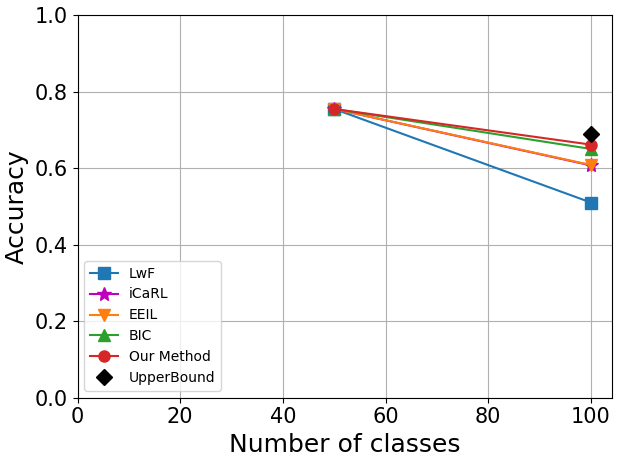}
    \parbox{15.5cm}{\small \hspace{2.cm}(d)}
\end{minipage}
\begin{center}
\vspace{-.5cm}
\caption{\textbf{Incremental learning results on CIFAR-100} with split of (a) 5 classes, (b) 10 classes, (c) 20 classes and (d) 50 classes. The \textbf{Upper Bound} in last step is obtained by offline training a model using all training samples from all classes. (Best viewed in color)}
\vspace{-1.1cm}
\label{result-cifar100}
\end{center}
\end{figure*}

\section{Experimental Results}
Our experimental results consists of two main parts. In part one, we compare our modified cross-distillation loss and the two-step learning technique as introduced in Section~\ref{method_part3} with current state-of-the-art incremental learning methods~\cite{EEIL,LWF,BiC,ICARL}. We follow the iCaRL experiment benchmark protocol~\cite{ICARL} to arrange classes and select exemplars, but in the more challenging online learning scenario as illustrated in Section~\ref{eval on CandI}. Our method is implemented on two public datasets: \textbf{CIFAR-100}~\cite{CIFAR} and \textbf{ImageNet-1000} (ILSVRC 2012)~\cite{IMAGENET1000}. Part two is designed to test the performance of our complete framework. Since our goal is to set up an incremental learning framework that can be applied to online learning scenario, we use \textbf{Food-101}~\cite{Food-101} food image dataset to evaluate our methods. For each part of our proposed framework, we compare our results to baseline methods as described in Section~\ref{Method}. 

\subsection{Datasets}
We used three public datasets. Two common datasets: CIFAR-100 and ImageNet-1000 (ILSVRC 2012) and one food image dataset: Food-101. 

\textbf{Food-101} is the largest real-world food recognition dataset consisting of 1k images per food classes collected from \textit{foodspotting.com}, comprising of 101 food classes. We divided $80\%$ for training and $20\%$ for testing for each class. 

\textbf{CIFAR-100} consists of 60k $32\times32$ RGB images for 100 common objects. The dataset is originally divided into 50K as training and 10k as testing.

\textbf{ImageNet-1000 (ILSVRC 2012)} ImageNet Large-Scale Visual Recognition Challenge 2012 (ILSVRC12) is an annual competition which uses a subset of ImageNet. This subset contains 1000 classes with more than 1k images per class. In total, there are about 1.2 million training data, 50k validation images, and 150k testing images. 

\textbf{Data pre-processing } For Food-101, we performed image resize and center crop. As for CIFAR-100, random cropping and horizontal flip was applied following the original implementation~\cite{RESNET}. For ImageNet, we follow the steps in VGG pre-processing~\cite{vgg}, including random cropping, horizontal flip, image resize and mean subtraction.
\vspace{-0.2cm}
\subsection{Implementation Detail}
Our implementation is based on Pytorch~\cite{pytorch}. For experiment part one, we follow the same experiment setting as current state-of-the-art incremental learning methods, a standard 18-layer ResNet for ImageNet-1000 and a 32-layer ResNet for CIFAR-100. For experiment part two, we applied a 18-layer ResNet to Food-101. The ResNet implementation follows the setting suggested in~\cite{RESNET}. We use stochastic gradient descent with learning rate of 0.1, weight decay of 0.0001 and momentum of 0.9. 

\textbf{Selection of block size $p$ in online learning scenario.} Different from offline learning scenario, where we select a batch size to maximize overall performance after many epochs. In online learning scenario, we need to select block size $p$ based on real life applications. More specifically, a large block size causes slow update since we have to wait until enough data arrives to update the model. On the other hand, using a very small block size, e.g., update with each new observation, although very fast, is not suitable for deep neural network due to strong bias towards new data. Therefore, we design a pretest using the first 128 new data for each experiment to repeatedly update the model by varying block size $p \in \{1,2,4,8,16,32,64\}$. Thus the optimal block size is chosen which gives the highest accuracy on these 128 new data. We do not consider $p > 64$ as such a large block size is not practical for real life applications. 

\begin{figure*}[htbp]
\centering
\begin{minipage}[t]{0.22\linewidth}
    \centering
    \includegraphics[width=4cm,height=3.5cm]{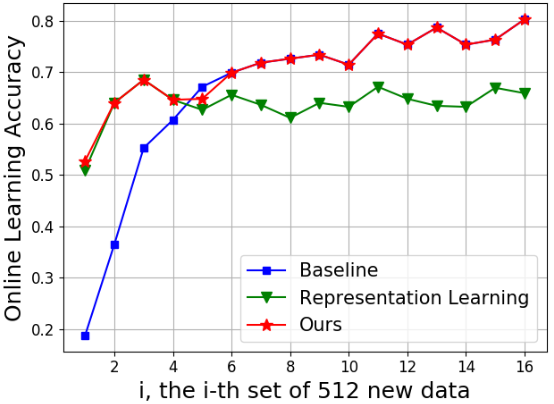}
    \parbox{15.5cm}{\small \hspace{2cm}(a)}
\end{minipage}
\hspace{2ex}
\begin{minipage}[t]{0.22\linewidth}
    \centering
    \includegraphics[width=4cm,height=3.5cm]{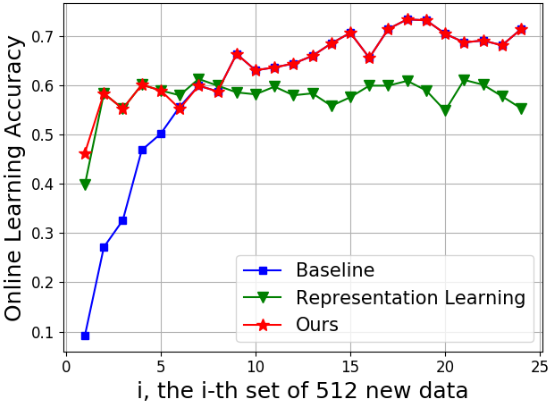}
    \parbox{15.5cm}{\small \hspace{2cm}(b)}
\end{minipage}
\hspace{2ex}
\begin{minipage}[t]{0.22\linewidth}
    \centering
    \includegraphics[width=4cm,height=3.5cm]{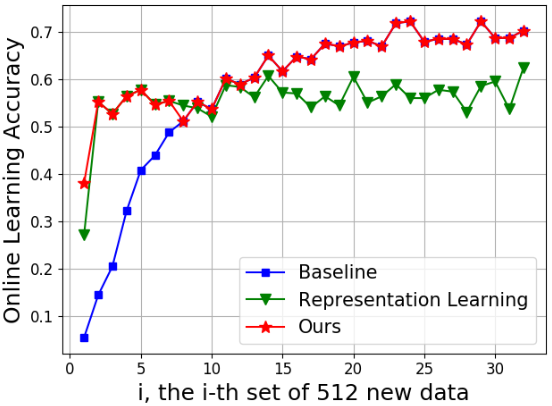}
    \parbox{15.5cm}{\small \hspace{2cm}(c)}
\end{minipage}
\hspace{2ex}
\begin{minipage}[t]{0.22\linewidth}
    \centering
    \includegraphics[width=4cm,height=3.5cm]{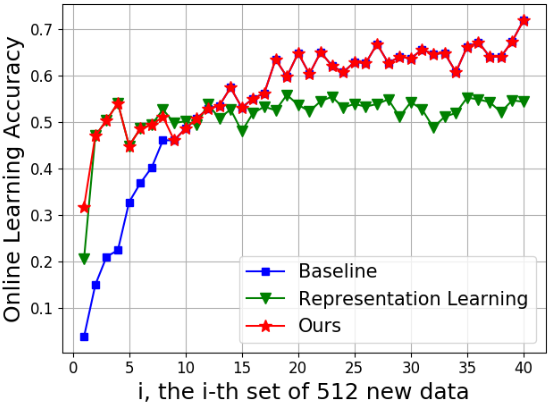}
    \parbox{15.5cm}{\small \hspace{2cm}(d)}
\end{minipage}
\begin{center}
\vspace{-0.5cm}
\caption{\textbf{Starting from scratch on Food-101} with number of new classes (a) 20 classes (b) 30 classes (c) 40 classes and (d) 50 classes. Baseline and our method are illustrated in Section~\ref{method_part1} (Best viewed in color)}
\vspace{-.8cm}
\label{result-online1}
\end{center}
\end{figure*}

\begin{table*}[!t]
\vspace{-0.2cm}
\begin{subfigure}{0.65\linewidth}
\centering
\begin{tabular}{|ccccc|}
\hline
 Method & 20 & 30 & 40 & 50\\
\hline\hline
Baseline & 62.81\% & 56.53\% & 54.35\% & 51.39\% \\
Representation Learning& 60.21\% & 55.32\% & 53.68\% & 51.26\% \\
Ours &\textbf{70.90\%} & \textbf{64.32\%} & \textbf{62.31\%} & \textbf{57.83\%} \\
\hline
\end{tabular}
\vspace{-0.15cm}
\caption{}\label{table:online1_online}
\end{subfigure}
\hspace{0.8cm}
\begin{subfigure}{0.35\linewidth}
\begin{tabular}{|c|c|c|}
\hline
  & Testing & Upper Bound\\
\hline\hline
20 & 78.77\% & 84.17\% \\
30 & 73.28\% & 80.95\% \\
40 & 71.42\% & 77.82\% \\
50 & 67.54\% & 74.46\% \\
\hline
\end{tabular}
\vspace{-0.15cm}
\caption{}\label{table:online1_test}
\end{subfigure}
\vspace{-0.4cm}
\caption{\textbf{Online learning from scratch on Food-101} with (a) Online accuracy and (b) Testing accuracy. The \textbf{Upper Bound} is obtained by offline training a model using all training samples from all given classes. (Best result marked in bold)}
\vspace{-.5cm}
\label{table_online1}
\end{table*}

\subsection{Evaluation of Modified Cross-Distillation Loss and Two-Step Learning}
\label{eval on CandI}
In this part, we compared our modified cross-distillation loss and two-step learning technique with the current state-of-the-art methods~\cite{ICARL,EEIL,BiC}. We consider the online setting that new classes data comes sequentially and we predict each new data at first and then use a block of new data to update the model. For each incremental step, we compare our accuracy obtained in online scenario with state-of-the-art results in offline mode. We constructed the exemplar set for both CIFAR and ImageNet with the same number of samples as in~\cite{ICARL,EEIL,BiC} for fair comparison. 
\vspace{-.2cm}
\begin{figure}[H]
\begin{center}
  \includegraphics[width=0.7\linewidth]{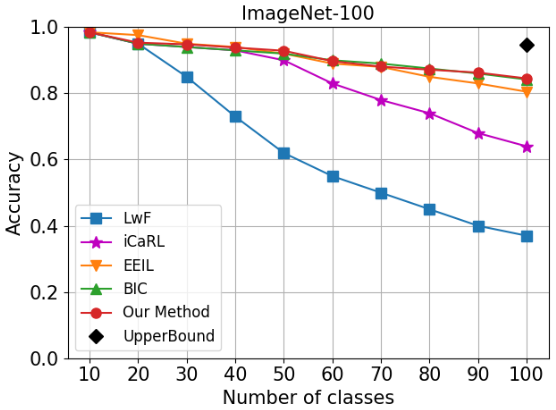}
  \vspace{-0.2cm}
  \caption{\textbf{Incremental learning results on ImageNet-100} with split of 10 classes. The 
  \textbf{Upper Bound} in last step is obtained by offline training a model using all training samples from all classes. (Best viewed in color)}
  \vspace{-.5cm}
  \label{fig:5}
\end{center}
\end{figure}
\vspace{-0.2cm}

\begin{table*}[htbp]
\begin{minipage}{1.0\linewidth}
\centering
\begin{tabular}{|c|c|c|c|c|}
\hline
\multicolumn{1}{|c|}{} & \multicolumn{2}{|c|}{Online Accuracy} & \multicolumn{2}{|c|}{Test Accuracy}\\
\hline
Incremental Step & new & old & new & old \\
\hline\hline
20 & 54.35\% $\rightarrow$ \textbf{64.78\%} & 22.83\% $\rightarrow$ \textbf{61.01\%} &\textbf{70.97\%} $\rightarrow$ 64.00\% &  41.77\% $\rightarrow$ \textbf{70.32\%} (84.17\%) \\
30 & 52.62\% $\rightarrow$ \textbf{62.25\%} & 22.41\% $\rightarrow$ \textbf{60.00\%} &\textbf{71.56\%} $\rightarrow$ 61.87\% &42.25\% $\rightarrow$ \textbf{69.90\%} (80.95\%) \\
40 & 46.30\% $\rightarrow$ \textbf{61.53\%} & 20.53\% $\rightarrow$ \textbf{53.43\%} &\textbf{66.62\%} $\rightarrow$ 56.31\% &40.82\% $\rightarrow$ \textbf{65.65\%} (77.82\%) \\
50 & 43.49\% $\rightarrow$ \textbf{56.76\%} & 19.47\% $\rightarrow$ \textbf{51.71\%} &\textbf{63.32\%} $\rightarrow$ 54.20\% &36.81\% $\rightarrow$ \textbf{63.92\%} (74.46\%) \\
\hline
\end{tabular}
\end{minipage}
\vspace{-0.2cm}
\caption{\textbf{Online learning from a trained model on Food-101} with \textbf{baseline method using original cross-distillation loss} in the left of $\rightarrow$ and \textbf{our proposed method} in the right (best result marked in bold), ($\sbullet[.5]$) shows the \textbf{Upper Bound} results.}
\vspace{-0.2cm}
\label{table_online3}
\end{table*}

\textbf{CIFAR-100.} We divided 100 classes into splits of 5, 10, 20, and 50 in random order. Therefore, we have incremental training steps for 20, 10, 5, and 2, respectively. The optimal block size is set as $p=8$. We ran the experiment for four trials and each time with a random order for the 100 classes. The average accuracy is shown in Figure~\ref{result-cifar100}. Our method shows the best accuracy for all incremental learning steps even in the challenging online learning scenario.

\textbf{ImageNet-1000.} As 1000-class is too large and impractical for online scenario, so we randomly selected 100 classes from the 1000 classes to construct a subset of the original dataset, which is referred to as ImageNet-100. We then divided the 100 classes into 10 classes split so we have an incremental step of 10. The optimal block size is set as $p=16$. We ran this for four trials as before and we recorded the average accuracy in each step as shown in Figure~\ref{fig:5}. Although the performance of EEIL~\cite{EEIL} surpass our method in the second step, we attain the best performance as more classes are added.

\subsection{Evaluation of Our Complete Framework}
\label{benchmark}

We used a food image dataset \textbf{Food-101}~\cite{Food-101} to evaluate performance of our proposed incremental learning framework. 

\textbf{Benchmark protocol of online incremental learning.} Until now, there is no benchmark protocol on how to evaluate an online incremental learning method. In addition to address catastrophic forgetting~\cite{CF} as in offline incremental learning, we also need to consider concept drift~\cite{CD} in online scenario. We propose the following evaluation procedure: for a given multi-class classification dataset, the classes should be randomly arranged. For each class, the training data should be further split into new training data and old training data. The former is used when a class is introduced to the model for the first time. The later is considered when the model has seen the class before, which is used to simulate real life applications and test the ability of the method to handle new observations of old classes.
After each online learning phase, the updated model is evaluated on test data containing all classes already been trained so far. There is no over-fitting since the test data is never used to update the model. In addition to the overall test accuracy, we should separately examine the accuracy for new classes and accuracy for old classes data. We also suggest to use online accuracy, which is the accuracy for data in training set before they are used to update the model, to represent the classification performance of future data stream. In general, online accuracy shows the model's ability to adapt to future data stream and online accuracy for old classes indicates the model's ability to handle new observations of old classes. 
\begin{figure*}[!t]
\centering
\hspace{.5cm}
\begin{subfigure}[t]{0.31\linewidth}
    \centering
    \includegraphics[width = 4.9 cm]{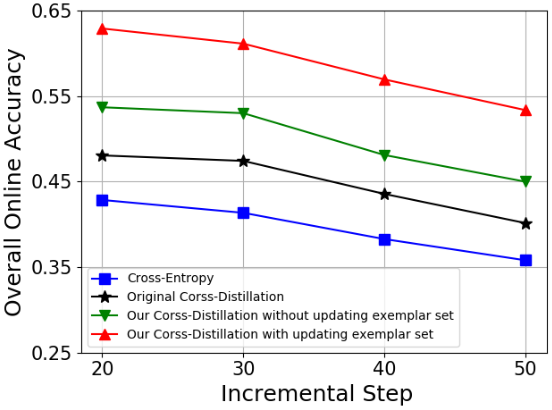}
    \vspace{-0.2cm}
    \caption{}\label{fig:ablation-a}
\end{subfigure}
\hspace{0.1cm}
\begin{subfigure}[t]{0.31\linewidth}
    \centering
    \includegraphics[width = 5. cm]{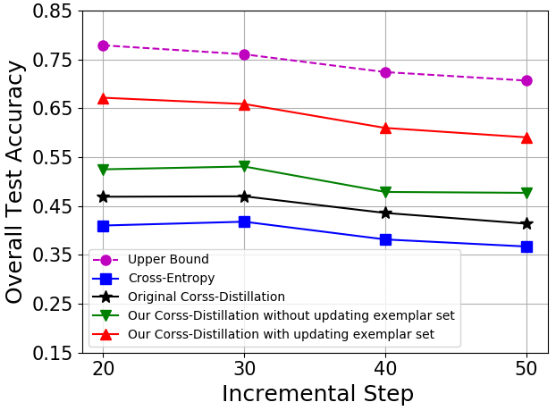}
    \vspace{-0.2cm}
    \caption{}\label{fig:ablation-b}
\end{subfigure}
\hspace{0.1cm}
\begin{subfigure}[t]{0.31\linewidth}
    \centering
    \includegraphics[width = 5. cm]{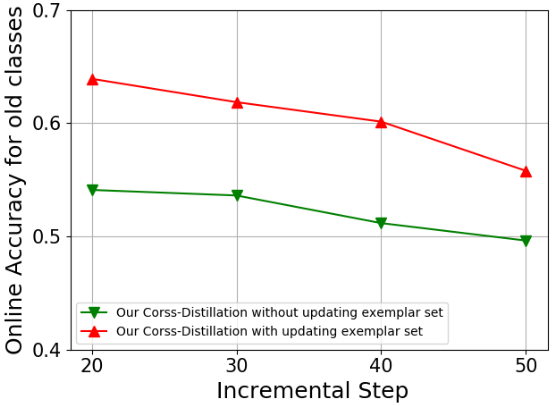}
    \vspace{-0.2cm}
    \caption{}\label{fig:ablation-c}
\end{subfigure}
\vspace{-0.3cm}
\begin{center}
\caption{\textbf{Ablation study on Food-101 dataset} (a) overall online accuracy (b) overall test accuracy (c) online accuracy for old classes. (Best viewed in color)}
\vspace{-.8cm}
\label{result-ablation}
\end{center}
\end{figure*}

\vspace{-0.2cm}
\subsection{Results on Food-101}
Although there are three separate components of the proposed incremental learning framework as described in Section~\ref{Method}, we only test the component described in~\ref{method_part1} once and then alternate between the two components described in~\ref{offline retraining} and~\ref{method_part3}. In addition, the offline retraining part in~\ref{offline retraining} is inapplicable with online incremental learning. So in this experiment, we test for one cycle of our proposed framework starting from scratch then learning from a trained model provided by offline retraining. We use half training data per class as new classes data and the other half as new observations of old classes. We divided the Food-101 dataset into split of 20, 30, 40, 50 classes randomly and performed the one incremental step learning with step size of 20, 30, 40, and 50, respectively. In addition, we construct exemplar set with only 10 samples per class to simulate real life applications instead of including more samples per class. 

\textbf{Learn from scratch.} In this part, we evaluate our method that combines baseline and representation learning as described in Section~\ref{method_part1}. Optimal block size is set as $p=16$. Result of online accuracy compared to baseline and representation learning is shown in Table~\ref{table:online1_online}. Our method achieved the best online accuracy in all incremental learning steps. Similarly, test accuracy compared to upper bound is shown in Table~\ref{table:online1_test}. We also calculated the accuracy of each 512 incoming new data as shown in Figure~\ref{result-online1}. We observed that the representation learning works well at the beginning when data is scarce and the baseline achieved higher accuracy as the number of new data increases. Thus by combining the two methods and automatically switch from one to the other, we attain a higher overall online accuracy. 

\textbf{Learn from a trained model.}
In this part, we perform a one incremental step experiment following our proposed benchmark protocol described in Section~\ref{benchmark} and the result is shown in Table~\ref{table_online3}. Compared to the baseline, our method improved the online learning accuracy for both new and old classes, which shows that our model can adapt quickly to future data stream including both new classes data or new observations of old classes. In addition, we significantly improved the test accuracy compared to the baseline method. However, the trade off is slightly lower accuracy for the new classes test accuracy compared to the baseline due to the use of the accommodation ratio in our method. 
Since it is difficult for the model to perform well on new classes without losing knowledge from the old classes, the accommodation ratio can be manually tuned to balance between the new classes and the old classes depending on the application scenario. A higher accommodation ratio leads to higher accuracy on new classes by trading off accuracy on old classes. For this experiment, we simply use $\beta = 0.5$.

\textbf{Ablation study.} We analyzed different components of our method to demonstrate their impacts. We first show the influence of different loss functions including cross-entropy, cross-distillation, and our modified cross-distillation. We then analyzed the impact of updating the exemplar set to mitigate concept drift. As shown in Figure~\ref{fig:ablation-a} and~\ref{fig:ablation-b}, even without updating exemplar set, our modified cross-distillation loss outperformed the other two (black and blue lines) for all incremental steps. By updating the exemplar set, we were able to achieve a higher overall online and test accuracy. Furthermore, Figure~\ref{fig:ablation-c} illustrates improvement of online accuracy for old classes by updating the exemplar set. Since we do not deliberately select any new data from old classes to update the model during the incremental learning step, as the data distribution changes, our method was able to automatically update the exemplar set by using the current class mean calculated by all data in old classes seen so far. Thus through the proposed two-step learning which pairs each new data with an exemplar, we can achieve a higher online accuracy for future data streams.

\section{Conclusion}
In this paper, we proposed an incremental learning framework including a modified cross-distillation loss together with a two-step learning technique to address catastrophic forgetting in the challenging online learning scenario, and a simple yet effective method to update the exemplar set using the feature of each new observation of old classes data to mitigate concept drift. Our method has the following properties: (1) can be trained using data streams including both new classes data and new observations of old classes in online scenario, (2) has good performance for both new and old classes on future data streams, (3) requires short run-time to update with limited data, (4) has potential to be used in lifelong learning that can handle unknown number of classes incrementally. Our method outperforms current state-of-the-art on CIFAR-100 and ImageNet-1000 (ILSVRC 2012) in the challenging online learning scenario. Finally, we showed our proposed framework can be applied to real life image classification problem by using Food-101 dataset as an example and observed significant improvement compared to baseline methods.
{\small
\bibliographystyle{ieee_fullname}
\bibliography{egbib}
}

\end{document}